\documentclass[letterpaper, 10 pt, conference]{ieeeconf}  
\IEEEoverridecommandlockouts       
\usepackage{cite}

\usepackage{amssymb}

\usepackage{xcolor,soul,framed} 

\colorlet{shadecolor}{yellow}

\usepackage[pdftex]{graphicx}
\graphicspath{{../pdf/}{../jpeg/}}
\DeclareGraphicsExtensions{.pdf,.jpeg,.png}

\usepackage[cmex10]{amsmath}
\usepackage{array}
\usepackage{mdwmath}
\usepackage{mdwtab}
\usepackage{eqparbox}
\usepackage{url}
\usepackage{multirow}
\usepackage[thinlines]{easytable}
\usepackage{comment}
\usepackage{makecell}

\title{\LARGE \bf
Soft Vision-Based Tactile-Enabled SixthFinger: Advancing Daily Objects Manipulation for Stroke Survivors} 

\author{Basma Hasanen $^{1,2}$,
Mashood M. Mohsan $^{1,2}$, Abdulaziz Y. Alkayas$^{1,2}$, Federico Renda $^{1,2}$
and Irfan Hussain $^{1,2}$
\thanks{Corresponding Author: Irfan Hussain, Email: irfan.hussain@ku.ac.ae
}
\thanks{ $^1$Center of Autonomous Robotics Systems, Khalifa University,  Abu Dhabi, United Arab Emirates, P O Box 127788, Abu Dhabi, UAE}
\thanks{$^2$Mechanical Engineering Department, Khalifa University,  Abu Dhabi, United Arab Emirates, P O Box 127788, Abu Dhabi, UAE}%
\thanks{ Corresponding Author: Irfan Hussain (irfan.hussain@ku.ac.ae)}
}

\providecommand{\keywords}[1]{\textbf{\textit{Index terms---}} #1}

\begin{document}
\maketitle
\thispagestyle{empty}
\pagestyle{empty}

\begin{abstract}
The presence of post-stroke grasping deficiencies highlights the critical need for the development and implementation of advanced compensatory strategies. This paper introduces a novel system to aid chronic stroke survivors through the development of a soft, vision-based, tactile-enabled extra robotic finger. By incorporating vision-based tactile sensing, the system autonomously adjusts grip force in response to slippage detection. This synergy not only ensures mechanical stability but also enriches tactile feedback, mimicking the dynamics of human-object interactions.
At the core of our approach is a transformer-based framework trained on a comprehensive tactile dataset encompassing objects with a wide range of morphological properties, including variations in shape, size, weight, texture, and hardness.
Furthermore, we validated the system's robustness in real-world applications, where it successfully manipulated various everyday objects. The promising results highlight the potential of this approach to improve the quality of life for stroke survivors.

\end{abstract}

\vspace{0.3cm}

\keywords Supernumerary Robotic Finger, Wearable Robots, Assistive Technologies, Tactile Sensing, Transformers
\endkeywords

\IEEEpeerreviewmaketitle

\section{Introduction}

Stroke frequently results in hand impairments and loss of grasping function. The devastating aftermath of a stroke often manifests in the form of hemiparesis—a debilitating unilateral weakness that disrupts daily life and compromises fine motor functions, especially those associated with grasping. While various rehabilitation devices exist to aid in hand function recovery after a stroke, only a small percentage of stroke patients fully recover hand functionality after six months \cite{nakayma1994compensation}. The prevailing compensatory aids are often single-purpose, bulky, and confined to structured environments, like rehabilitation centers \cite{hussain2017toward}. 

\begin{figure*}
    \centering
    \includegraphics[width=1.8\columnwidth]{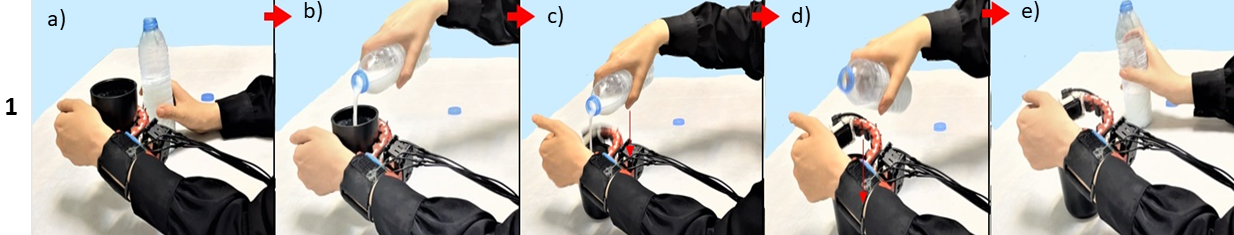}
    \caption{The figure depicts a grasping failure scenario during liquid pouring when both functional and non-functional hands are used and force is not auto-regulated.}
    \label{fig:problem}
    \vspace{-0.4cm}
\end{figure*}

Assistive robotic devices designed for stroke survivors emerge as a beacon of hope, epitomizing the potential of technology in medical compensation \cite{hussain2017toward, hussain2017soft, hussain2020compensating, hasanen2022novel, hendriks2024enhancing}.
In \cite{hussain2016soft}, a supernumerary robotic finger, called the SixthFinger, has been proposed to compensate for the missing grasping abilities in hemiparetic upper limb. Unlike traditional assistive devices that often serve a singular, focused purpose, this robotic finger is designed to be an all-encompassing solution. 
It can assist post-stroke patients to overcome the multifaceted challenges in daily bimanual tasks \cite{hussain2017toward}—challenges that are not only physical but also psychological in nature \cite{carota2005psychopathologie}. Based on tests with real patients, providing an extra finger can potentially build confidence and a renewed sense of independence in the user \cite{hussain2016soft}. Different control strategies has been proposed for the device. In \cite {hussain2015using}, a trigger-based control technique was used to control the SixthFinger. Although this strategy is straightforward and intuitive, the control interface involves the human hand thumb, restricting the use of the thumb in task completion. 
In \cite{hussain2016emg}, another EMG electrodes interface has been developed to make the device's control more intuitive.  
Nevertheless, the user is required to utilize their non-paretic arm to provide gestures for regulating the gripping force, a task that becomes impractical if the non-paretic arm is already engaged with the grabbed object. When the patient's non-paretic hand is busy, they might struggle to maintain consistent manual control over the device's exerted force. This inconsistency can lead to problems: insufficient force might cause the grasped object to slip, as illustrated in Fig. \ref{fig:problem}. Additionally, the manual tuning of the grasping force increases the cognitive load of the patient. 

\begin{figure}
    \centering
    \includegraphics[width=3.5in]{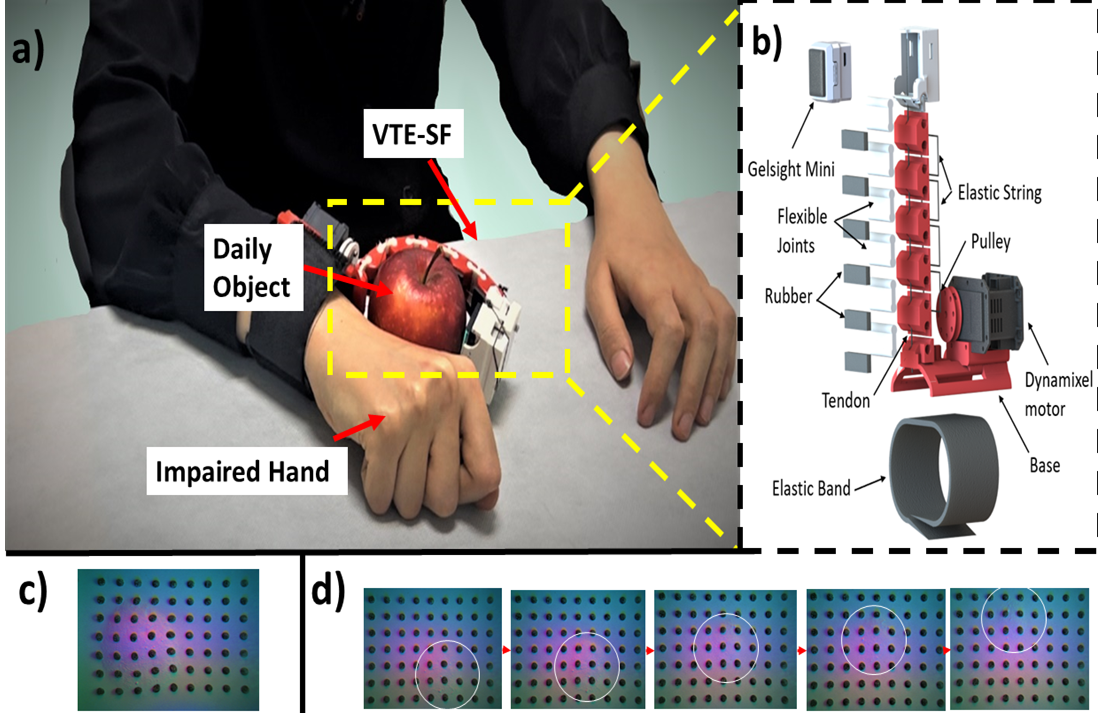}
    \caption{(a) The VTE-SF system assists the stroke survivor to grasp an object (an apple) (b) The CAD design of VTE-SF. (c) The image captured by GelSight mini when the device touches the object. (d) The Sequence of frames captured by Gelsight mini when slippage of the object occurs.}
    \label{fig:firstfig}
    \vspace{-0.5cm}
\end{figure}

In this work, to address the limitations mentioned above, the design of the SixthFinger is enhanced by adding a tactile sensor in the fingertip. 
The inclusion of tactile sensing for grasping simulates the natural feedback loop of a biological finger: when slip is detected, grip force increases in human grabbing, according to early research \cite{westling1984factors}. 
Commercial tactile sensors exhibit deficiencies in multiple aspects. Typically, the measurement focus of these sensors lies on normal force or pressure, rather than shear force, hence disregarding the tangential forces. The determination of incipient slip or entire slip is challenging in the absence of shear field measurements and the observation of its temporal evolution.
Vision-based tactile sensors, however, expand this horizon by utilizing visual cues to interpret tactile interactions. Through integrated cameras and advanced algorithms, these sensors can provide a more detailed understanding of the interaction between the manipulator and the object in contact, allowing for better grip adjustments and more precise interactions.  
By providing real-time feedback on grasp stability, it not only ensures the safety of the object being held but also instills confidence in the user.

Our contributions in this paper lie in: (1) A novel, soft, Vision-Based Tactile-Enabled SixthFinger system (VTE-SF).
(2) A transformer-based grasping framework for object grabbing that can detect the slippage and modulate gripping force using tactile images. We conducted ablation studies to enhance the performance and accuracy of the proposed approach. (3) In addition, we experimentally tested the proposed system in handling different daily objects using the VTE-SF device and demonstrated its practicality.

\section{Related Work}
\label{section two}

\subsection{The Gelsight Vision-based Tactile Sensor}

GelSight is a vision-based tactile sensing modality developed by \cite{yuan2017gelsight}. It is made up of four primary parts \cite{jia2013lump}: 1) a transparent elastomer piece with a reflective surface on one side, 2) a clear supporting plate made of either glass or acrylic for this elastomer, 3) consistent lighting typically supplied by LEDs, and 4) a camera positioned behind the support plate to record the image imprinted on the elastomer. Upon engagement with an object's surface, the elastomer molds in a manner that reflects the topography of the interfaced surface. Such deformations, when illuminated, are optically captured using the camera. Gelsight mini sensor is used in this paper. Distinguished by its high spatial resolution (3280 x 2464 pixels), GelSight mini is synergetic with computer vision models. 
Further enhancing its utility is the soft contact surface, ensuring minimal invasiveness during interactions, and making it ideal for delicate tasks. Notably, despite its cutting-edge capabilities, GelSight remains a cost-effective solution.

\subsection{Deep learning for slippage detection in grasping  and tactile perception}

Slip, often resulting from insufficient grip strength or incorrect grip placement, has been a subject of extensive research \cite{chen2018tactile}. Such slips indicate unstable holding during robotic manipulation. By recognizing or anticipating these slips, robots can adjust their grip technique and strength to ensure successful grasping. This significance has led to the development of numerous sensors \cite{francomano2013artificial}. Understanding these contact dynamics is crucial for effective robotic handling. As computer vision progresses and with the advent of optical tactile sensors, there's a renewed approach to how tactile images are processed. By synergizing tactile data with vision-based techniques, enhanced outcomes have been observed.  Calandra et al. in \cite{calandra2017feeling, calandra2018more} utilizes a singular visual and tactile image. This approach anticipates grasp stability and provides guidance for potential re-grasping before lifting. \cite{li2018slip} presents a novel approach utilizing deep neural networks (DNN) for the purpose of slip detection. 
\cite{yuan2015measurement} proposed the tracking-surface-marker technique. This method allows the GelSight elastomer surface movement to show the external contact pressure, effectively signaling partial slip during shear loading.
In \cite{yan2022detection}, A unique CNN-TCN model was introduced to merge tactile and visual data for identifying the beginning or continuation of slip. The model they suggested attains a detection accuracy of 88.75\% and surpasses the CNN-LSTM model paired with various pre-trained vision networks. Li et al. \cite{li2019rotational}, utilizes a sequence of images from an external camera and an optical tactile sensor to discern the direction of rotation as an object is lifted. 
In \cite{han2021learning}, the Transformer models were assessed on a public dataset specifically for slip detection. Their results highlight that, in terms of accuracy and computational efficiency, these models exceed the performance of the CNN+LSTM model. The superiority of combining visual and tactile learning over using a singular modality has been highlighted in previous studies, which delved into various visual-tactile learning tasks \cite{rouhafzay2020transfer, lin2019learning, lee2019touching}.

\begin{figure*}[htp!]
    \centering
    \includegraphics[width=2.1\columnwidth]{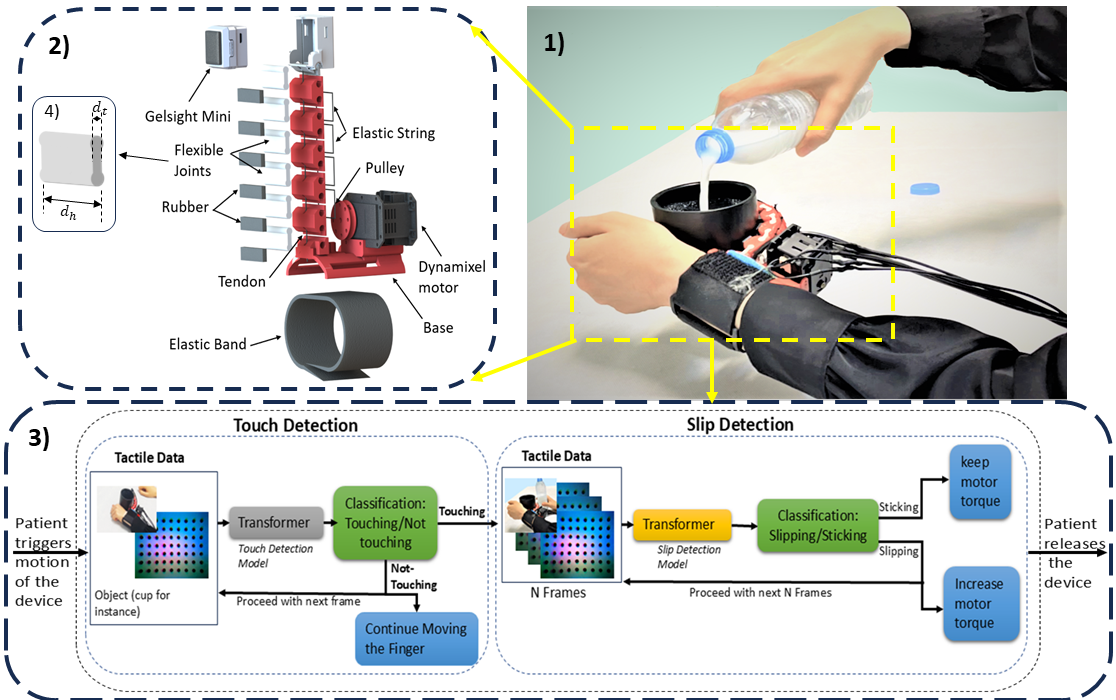}
    \caption{ Proposed System Overview: 1) The VTE-SF system supporting object grasping for patient. 2) The CAD model of the proposed device. 3) System Operation Sequence: Patient positions hand near the object, initiating device motion $\rightarrow$ Touch model detects device-object contact $\rightarrow$ On touch, slip model monitors for slippage $\rightarrow$ Grasping force auto-adjusts if slippage occurs $\rightarrow$ After task completion, patient commands finger release. 4) The dimensions of the soft joint are labeled: $d_h$, and $d_t$ are the height and thickness, respectively.
    }
    \label{fig:frame}
    \vspace{-0.3cm}
\end{figure*}

\section{Solution Overview}

In this paper, we present an adaptive robotic finger system to assist stroke survivors. This innovative solution employs the patient's paretic limb as the static element of a gripping mechanism. Concurrently, the robotic extra finger is posited as the dynamic counterpart, collaboratively facilitating an effective grasping mechanism. To operationalize this device, a systematic grasping framework has been proposed. The complete system is illustrated in Fig. \ref{fig:frame}. The sequence of operation begins with the patient initiating the robotic finger's motion. Upon activation, the system utilizes a touch detection model for contact detection. Once the touch between the object and the device is identified, potential object slippage is ascertained using a slippage detection model. In instances of detected slippage, the system independently calibrates the exerted force, increasing it until secure grip is established and slippage is stopped. Integrating the touch and slip detection models has augmented the proficiency of the grasping approach. Upon task completion, patients can command the release of the finger. This methodology underscores the integration of human intention with robotic precision, offering a tailored solution to the unique challenges presented by hemiplegia.

\begin{figure}
    \centering    \includegraphics[width=0.8\columnwidth]{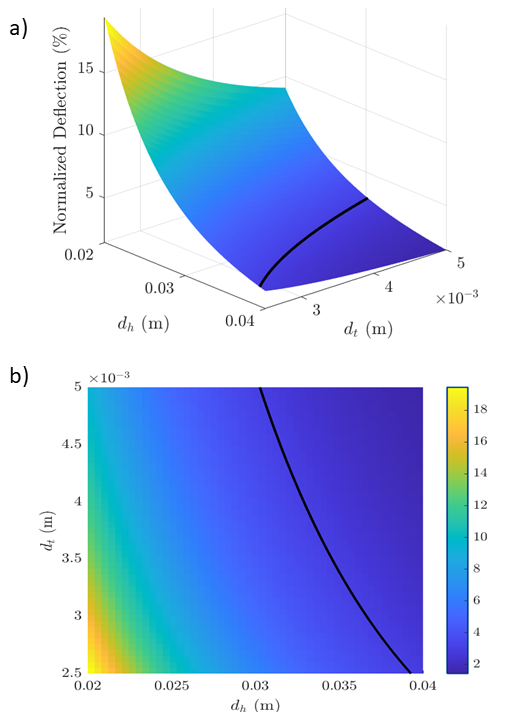}
    \caption{Results from the parametric study conducted using the SoRoSim MATLAB toolbox: a) The 3D surface, b) The x, y projection: $d_h$, and $d_t$ are the height and thickness of the soft joints, respectively. }
    \label{fig:Deflection}
    \vspace{-0.5cm}
\end{figure}

\section{Proposed Approach}
\label{section three}

\subsection{Design and Fabrication}
\subsubsection{The Vision-based Tactile Enabled SixthFinger (VTE-SF)}

In this paper, the proposed prototype, seen in Fig. \ref{fig:frame} (2), adheres to the same operational principle seen in \cite{hussain2016emg}. As depicted in Fig. \ref{fig:firstfig}, it compensates for hand grasping by opposing the paretic limb with the extra robotic finger. However, the fingertip is modified to attach the Gelsight mini  sensor. 
The device is composed of a modular flexible finger and a supporting base. A combination of 3D printed PLA polymer for rigidity and thermoplastic polyurethane for flexibility is used. 
The device's movement is powered by a single actuator, which operates a tendon or fishing wire that runs through the rigid section. This tendon's design, with one end attached to the fingertip and the other to a pulley on the actuator, guarantees grip stability. Moreover, the inherent design enhances the grip's efficiency by automatically adjusting to uncertainties.
The device's actuation is driven by a Dynamixel servo MX-64 from Robotis, South Korea. To control the (Dynamixel MX-64), the U2D2 board and the U2D2 power hub were used, interfacing through a 3Pin TTL connector. 

Due to the soft nature of the flexible joints, the finger is prone to noticeable bending under its own weight, especially with the presence of the tactile sensor at the tip. Thus, in order to account for that, we conduct a parametric study to guide our design choices using the MATLAB toolbox SoRoSim \cite{mathew2022sorosim}, \cite{mathew2024reduced} which simulates soft, rigid and hybrid systems. In this study, we vary the thickness and height of the soft joints ($d_t$ and $d_h$, respectively, in Fig. \ref{fig:frame} (4)) while monitoring the tip’s out-of-plane deflection due to the finger's weight. We aim for a deflection not exceeding 3$\%$ of the total length of the finger. Fig. \ref{fig:Deflection} shows how the normalized deflection vary with each parameter, and the black curve being the iso-line for 3$\%$. By setting the height equal to that of the tactile sensor, the corresponding thickness required to achieve the allowable deflection can be identified. Thus, for a height of 3.4 cm, the flexible joint thickness should be 3.8 mm.

\begin{figure}
    \centering
    \includegraphics[width=0.9\columnwidth]{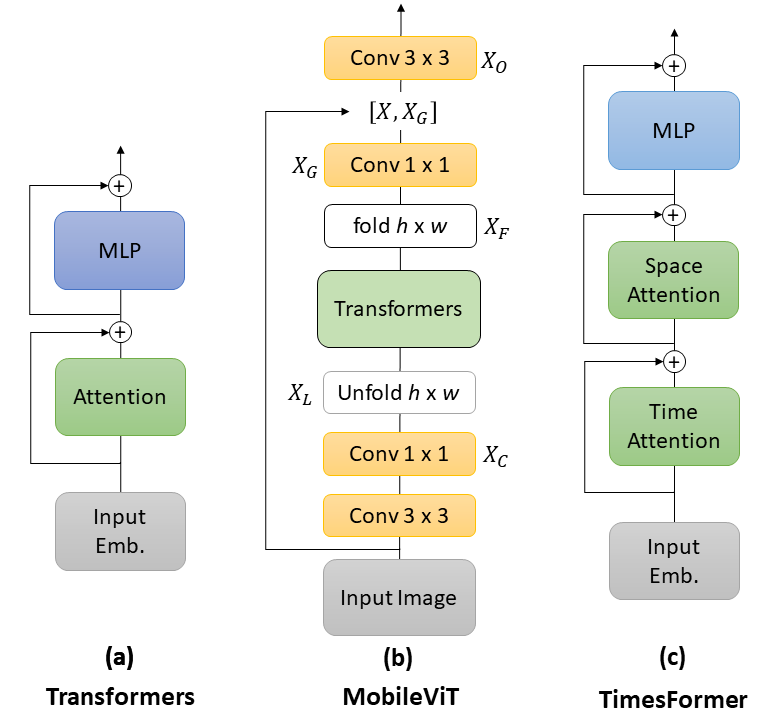}
    \caption{Transformers architecture}
    \label{fig:transformer}
    \vspace{-0.5cm}
\end{figure}

\subsection{Vision Transformers for touch and slip detection}
Utilizing transformer architectures has emerged as a novel and promising approach for challenging problems in the robotic domain. Transformers \cite{han2022survey} function as a sequential model. They process a series of inputs, such as image patches. Each input patch undergoes embedding through linear projections. These embedded inputs are then passed to self-attention layers, which focus on each patch. After that, the attended patches are normalized using layer normalization. Subsequen/tly, the output is directed to a Multi-Layer Perceptron (MLP) in a bottleneck fashion to derive global features. The central component of a Transformer model is its self-attention mechanism, which identifies and focuses on crucial features from the input. Once the embedding vector is combined with positional encoding, the result is channeled through multiple linear layers to produce Q (Query), K (Key), and V (Value) vectors for self-attention computation. The subsequent equation depicts the calculation for the attention mechanism:

\begin{equation}\label{eq_2}
\text{Self Attention }(Q,K,V) = \text{softmax}((Q \cdot K^{\text{T}})/\sqrt{d_{k}})\cdot V
\end{equation}

In Transformers, the dot product of \textit{Q} and \textit{K} vectors streamlines computations, enabling the creation of multi-headed self-attention (MSA) layers. Subsequently, the softmax function is used to allocate probabilities to the most significant values from the \textit{V} vector. 

\subsubsection{MobileViT for touch detection} The MobileVit architecture was selected for its lightweight structure, making it an ideal choice for the relatively low-complexity task of touch detection.
MobileViT, introduced in \cite{mehta2021mobilevit}, offers a specialized vision transformer optimized for mobile devices by seamlessly integrating the strengths of CNNs and Transformers. This design, illustrated in Fig. \ref{fig:transformer} (b), focuses on efficient representation of both localized and broad data from input tensors. It emphasizes maintaining a receptive area of $H \times W$ and captures an image's spatial context and attributes.
The core approach with MobileViT is to unfold the tensor $X_L$ into $N$ distinct flattened patches $X_U$, where $P = wh, N = HW$ and $P$ denotes the patch count. The process encodes inter-patch dynamics with transformers to derive $X_G$:
\begin{equation}\label{eq_5}
\text{X$_G$}(p) = \text{Transformers}(X_{U}(p)), 1 \leq p \leq P
\end{equation}
Differing from conventional vision transformers, MobileViT preserves both patch and intra-patch spatial arrangements. After refining 
$X_G$ to 
$X_F$, it's subsequently projected to a reduced 
$C-dimensional$ space and merged with $X$. The result is blended using an 
$n \times n$ convolutional layer. Given that both local and global data are encoded, each pixel in $X_G$ holds information from the entirety of $X$, achieving an effective receptive field of 
$H \times W$.
Conceptually, MobileViT can be viewed as a fusion of transformer and convolutional techniques, optimized for compatibility and efficiency across varied computational environments without the need for additional modifications.

\begin{figure*}
    \centering
    \includegraphics[width=2\columnwidth]{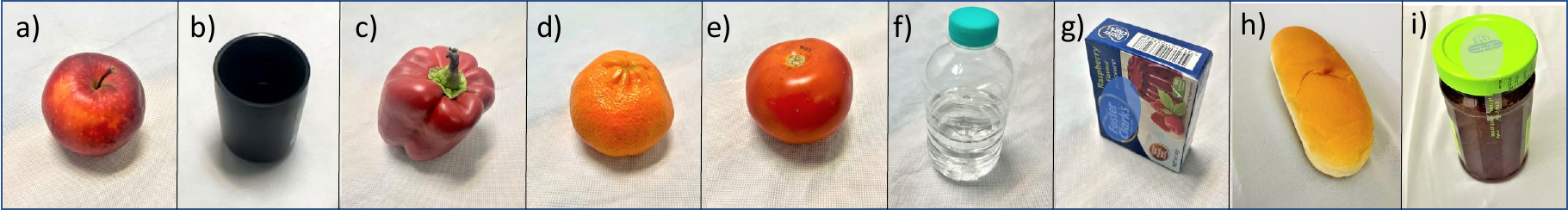}
    \caption{Objects in our dataset: a) Real apple, b) Cup, c) Real capsicum, d) Real orange, e) Real tomato, f) Bottle, g) Gelatin box, h) Bread, i) Jam Jar.}
    \label{fig:items}
    \vspace{-0.3cm}
\end{figure*}

\begin{figure}
    \centering
    \includegraphics[width=1\columnwidth]{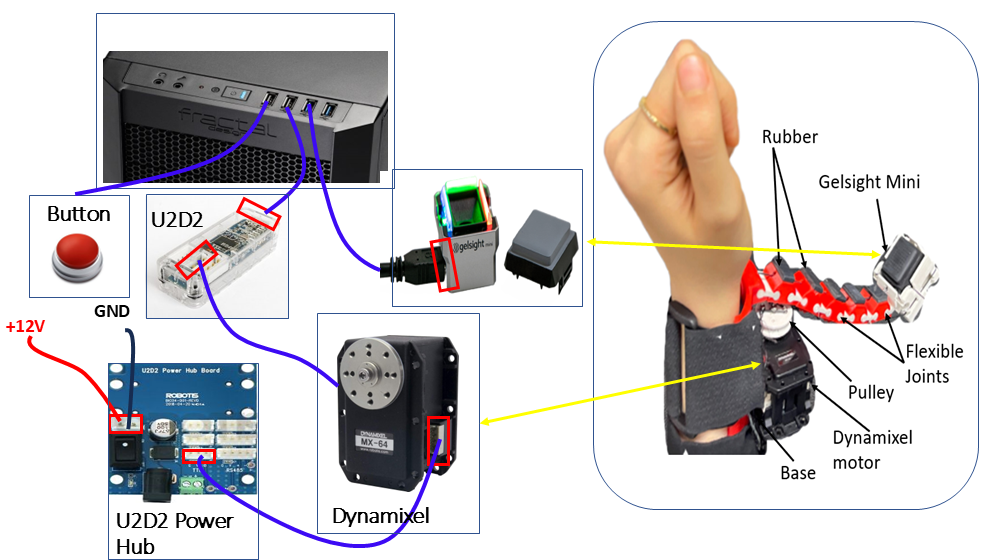}
    \caption{The Experimental Setup: the main components are highlighted}
    \label{fig:Expset}
    \vspace{-0.3cm}
\end{figure}

\subsubsection{TimeSformer for slip detection}
Transformers have recently gained substantial interest. The foundational Vision Transformer (ViT) \cite{dosovitskiy2020vit} adapted transformers originally developed for text classification to perform image classification. The TimeSformer model, first introduced in \cite{bertasius2021space}, addresses video processing by handling temporal and spatial attention separately, thereby reducing overall computational complexity.
It processes $F$ RGB frames of dimensions $H \times W$. These frames are broken down into $N$ distinct patches of size $P \times P$ and then linearized into vectors. The key equation that defines the temporal attention within each block is depicted in Eq. \ref{eq_7}:

\begin{equation}\label{eq_7}
{\alpha^{(l,a)time}_{(p,t)}} = SM\left(\frac{{q^{(l,a)}_{(p,t)}}^T}{\sqrt{D_h}} \left[ k^{(l,a)}_{(0,0)} \left\{ k^{l,a}_{p',t'} \right\}_{\substack{p = 1,..,N \\ t = 1,..,F}}  \right]  \right)
\end{equation}

Following the temporal attention process, the resulting encoding 'z' is employed for spatial attention computation. A noteworthy aspect of this method is its efficiency in matrix learning, requiring only (N+F+2) comparisons per patch.
The main block of TimeSformer is shown in Fig. \ref{fig:transformer} (c).

\subsection{Dataset and Data Collection}
To develop a refined slip detection model, we utilized the dataset from \cite{li2018slip}. Given the limited size of this dataset, we augmented it by gathering our own dataset using a selection of 9 everyday objects, as depicted in Fig. \ref{fig:items}. These objects, which vary in form, weight, compliance, and friction coefficients, were chosen from the Yale-CMU-Berkeley (YCB) Object and Model set. This set, introduced in \cite{calli2015benchmarking}, was designed to standardize benchmarking in robotic manipulation, rehabilitation research, and prosthetics, encompassing commonly used objects in manipulation tests. Our slip data collection approach was inspired by \cite{li2018slip}. Furthermore, using the objects in Fig. \ref{fig:items}, we gathered data to train and validate the used touch detection model. 
\subsection{Ablation Studies}

For a more profound understanding of the baseline slip detection model, several ablation studies were conducted. This involved modifying certain architectural elements to improve accuracy while simultaneously simplifying the model to reduce computational demands. The outcomes of these ablation studies can be found in Table. \ref{tab:ablation}.

\begin{figure*}[htp!]
    \centering
    \includegraphics[width=1.9\columnwidth]{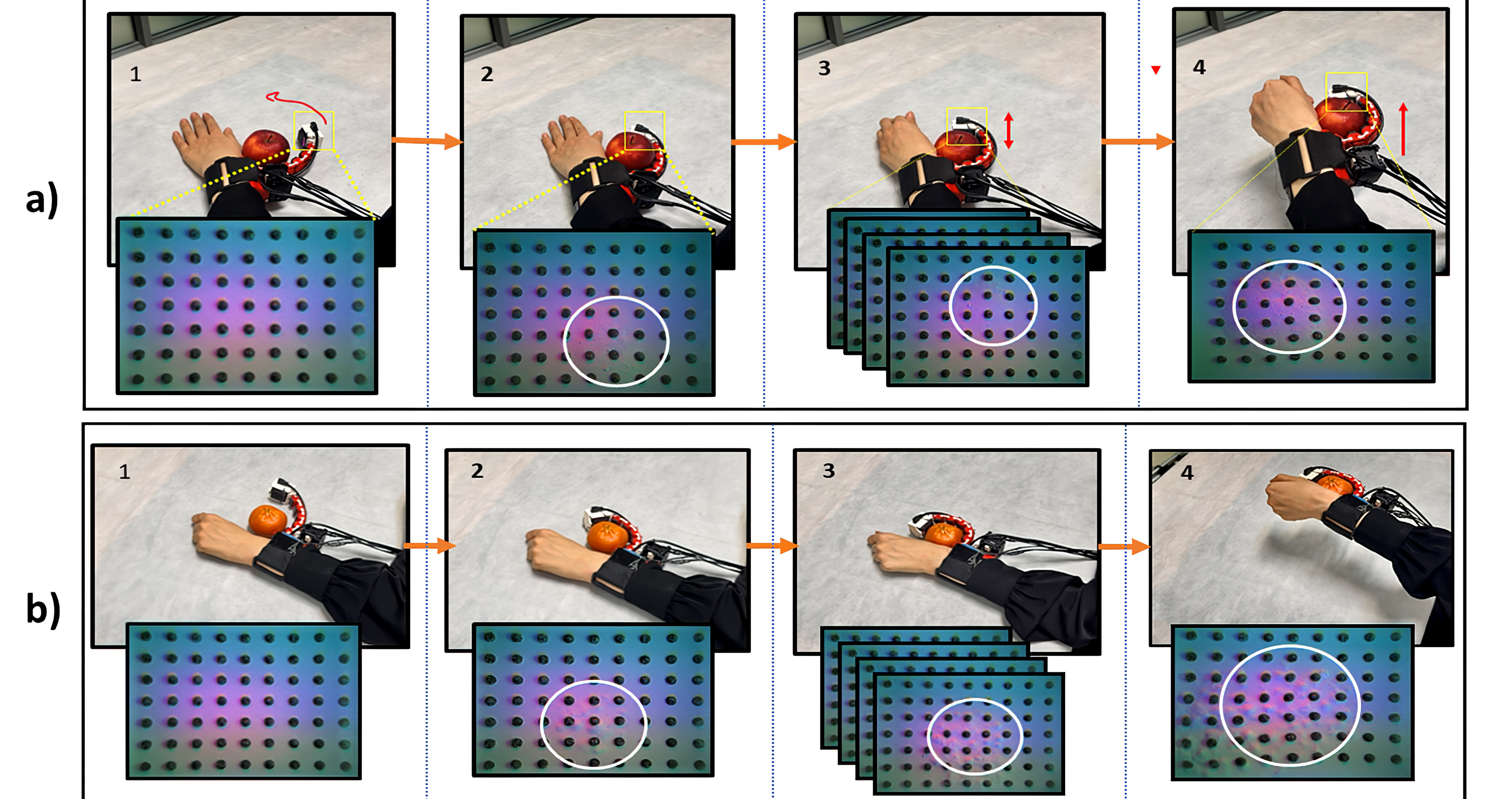}
    \caption{The figure demonstrates the grasping procedure using the proposed device and approach. The tactile images are shown in the figure's bottom half: (1) VTE-SF starts, initiated by the patient. (2) VTE-SF approaches and makes contact with the object. (3) The user raises their arm, sliding the device's tip on the object. (4) The proposed model detects slips and auto-regulates the grasping force, enabling secure object lifting.}
    \label{fig:ExpPro}
    \vspace{-0.3cm}
\end{figure*}

The following is a breakdown of each ablation study:

\subsubsection{AB-1 (Hidden sizes)} 
In the slip detection model, the "hidden size" represents the dimensionality of the model's internal vector representations of input segments. It's pivotal for the model's computational demands and its capacity to capture patterns. To strike a balance between efficiency and accuracy, we modified the baseline slip detection model's hidden dimension from 768 to 384 and 576. However, this led to a considerable 10\% decline in model accuracy, underscoring the importance of this hyperparameter.

\subsubsection{AB-2 (Attention heads)}Our experimentation involved modifying the number of self-attention heads in the slip detection model by a multiplicative factor of 4. While amplifying the number of attention heads enhances model accuracy, it also increases the model's complexity. This presents a computational challenge, especially when computing attention in video input, resulting in a suboptimal trade-off.

\subsubsection{AB-3 (Blocks count)} While the baseline model encompasses 12 blocks, we experimented by reducing this to 8, 6, and 4 blocks. Notably, when the model was adjusted to 8 blocks, there was an enhancement in accuracy. The primary component of the slip detection model, as illustrated in figure \ref{fig:transformer} (a), shows that an increase in the number of blocks inherently amplifies complexity of the model.

\begin{table}[t]
\footnotesize
    \centering
    \caption{Ablation studies results.
    }
    \begin{tabular}{c|c|ccc|c}
         \hline 
         Sr & Name & \makecell{Hidden\\ size} & \makecell{Attention \\heads} & \makecell{Encoder \\blocks} & Accuracy
         \\\hline
        1& Baseline & 768 & 12 & 12 & \underline{0.8615}\\
        \hline
        2& \multirow{2}{*}{\textit{AB-1}} & 384 & 12 & 12 & 0.7307 \\
        3& & 576 & 12 & 12 & 0.7076 \\
        \hline
        4& \multirow{2}{*}{\textit{AB-2}} & 768 & 16 & 12 & \textbf{0.8923} \\
        5& & 768 & 8 & 12 & 0.7923 \\
        \hline
        6& \multirow{3}{*}{\textit{AB-3}} & 768 & 12 & 8 & \textbf{0.8923} \\
        7& & 768 & 12 & 6 & 0.8615 \\
        8& & 768 & 12 & 4 & 0.8076 \\        
        \hline
    \end{tabular}
    \label{tab:ablation}
\end{table}

\begin{table}[t]
\footnotesize
    \centering
    \caption{Comparison with other models.
    }
    \begin{tabular}{c|c|c|c}
         \hline 
         Sr & Name & Dataset & Accuracy \\\hline
         1 & CNN + LSTM & \multirow{6}{*}{\textit{Slip Detection \cite{li2018slip}}} & 0.806 \\
         2 & ViVit & & 0.818 \\
         3 & TimesFormer & & 0.81 \\
         4 & X-clip & & 0.7384 \\
         5 & Ours (baseline)& & 0.8615 \\
         6 & Ours (proposed) & & 0.8923 \\
         \hline
         7 & Ours (proposed) & \textit{Our dataset} & 0.85 \\
        \hline
    \end{tabular}
    \label{tab:compare}
    \vspace{-5mm}
\end{table}

\begin{table}[t]
\footnotesize
    \centering
    \caption{List of hyperparameters and model configuration.
    }
    \begin{tabular}{c|c|c|c}
         \hline 
         Sr & Parameter / Configuration & Touch model & Slip model \\\hline
        1 & Number of frames & 1 & 8 \\
        2 & Image size & 256 & 224 \\
        3 & Number of Channels & 3 & 3 \\
        4 & Hidden dimension/s & 144, 192, 240 & 768 \\
        5 & Activation Function & SiLU & GELU\\
        6 & Layer normalization & - & 1e-06\\
        7 & Intermediate size& - & 3078 \\
        8 & MLP ratio & 2.0 & - \\
        9 & Patch size & - & 16 \\
        \hline
    \end{tabular}
    \label{tab:para}
\end{table}

\subsection{Comparison of our model with other models}
Upon finalizing the slip detection model through ablation studies, a comparative analysis against  state-of-the-art methods was undertaken. As shown in Table \ref{tab:compare}, our model was compared with the CNN + LSTM model from \cite{li2018slip}, ViVit model from \cite{arnab2021vivit}, TimesFormer model from \cite{bertasius2021space}, and X-clip model from \cite{ma2022x}. All evaluations were derived from fine-tuning the respective models on both the existing slip dataset and our own. The results highlight that our proposed model achieves competitive performance metrics relative to its contemporaries. 

It should be highlighted that the used touch detection model did not undergo ablation studies, because the MobileViT architecture, when fine-tuned on our dataset, approached a touch detection accuracy of nearly 100\%.

\section{Experiments and Results}
\label{section four}

\subsection{Experimental Setup and Implementation Details}

Figure \ref{fig:Expset} depicts the experimental setup employed to assess the proposed device, illustrating the connections between all components. This setup comprises a PC, power hub, U2D2 communication module, button, and the proposed system. The button functions as a trigger to initiate and end motion, while the power hub supplies a 12V input to the Dynamixel motor for actuation. Each connection is shown to provide a clear representation of the system’s integration and operational framework.

In terms of training specifics, our proposed grasping framework is developed using PyTorch and undergoes training on a GeForce RTX 4090 24GB GPU. Both models utilize pre-trained weights for fine-tuning on our touch dataset, slip dataset from \cite{li2018slip}, and our slip dataset (collected using the VTE-SF device) throughout the training process. The networks are refined using the AdamW optimizer over 10 epochs with batch sizes of 16 and 32. During evaluation, we adopt the parameter settings that yield optimal outcomes on the validation set. A comprehensive overview of hyperparameters and the model's setup can be found in Table \ref{tab:para}.
\begin{figure*}[htp!]
    \centering
    \includegraphics[width=1.8\columnwidth]{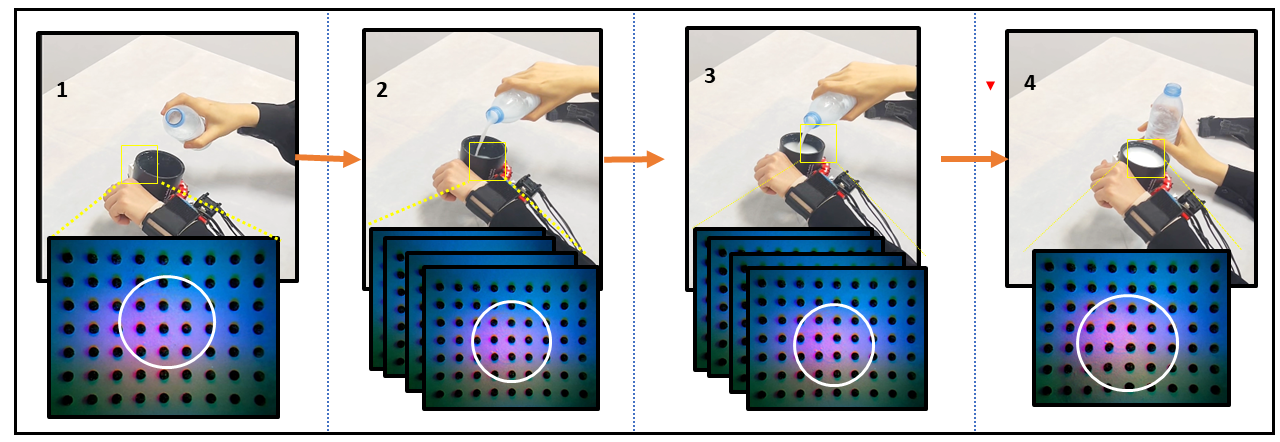}
    \caption{ The figure presents a sequence emphasizing slippage incidents due to interactions between the object and the patient’s non-paretic hand. The sequence includes: (1) Initial grasp of the object; (2), (3) Progressive increase in object weight through fluid addition; (4) Final grasp of the object}
    \label{fig:ExpPro2}
    \vspace{-0.3cm}
\end{figure*}

\subsection{Demonstration of the proposed approach}

We deployed the trained models on the VTE-SF device to demonstrate the grasping system in two contexts: managing daily items with the non-paretic hand and maintaining a grip when the patient's non-paretic hand interacts with the held object (illustrated in Fig. \ref{fig:problem}). 

One demonstration sequence is shown in Fig. \ref{fig:ExpPro}, during the demonstration, we evaluated the device's adeptness in initially sliding over, then firmly holding various daily objects. This sequence, where the robotic finger skims the object's surface before securing a grip, enables both the hand and object to move cohesively. Our objective was to ascertain the success rate and reliability across both seen (Fig. \ref{fig:ExpPro} (b)) and unseen objects (Fig. \ref{fig:ExpPro} (a)). For both seen and unseen objects (Shown in Fig.\ref{fig:items}), the human performed 30 grasp attempts for each object. The system achieved grasping success rates of 100\% for both seen and unseen objects, with each attempt completed in less than 20 seconds. 

Fig. \ref{fig:ExpPro2} presents another demonstration sequence, emphasizing slippage incidents resulting from interactions between the patient’s non-paretic hand and the object. This test aimed to evaluate the device’s autonomous response to such challenges. To simulate real-world variations, the object's weight was incrementally adjusted by adding fluid to the grasped item. This task was repeated 30 times, yielding a success rate of 90\%.

\section{Conclusion}
\label{section five}

In this paper, we introduce a novel transformer-based grasping method to assist stroke survivors. This approach employs an innovative assistive soft robotic finger, equipped with a vision-tactile sensor at its tip. This design aims to ensure a secure and safe grasp, and to diminish the stroke survivor's cognitive load while encouraging the use of their residual mobility. Our method's efficacy has been validated through practical evaluations across various objects and some potential applications has been demonstrated.

In the future, we plan to test the device with patients through user studies. This will involve exploring other potential applications, interfaces and haptic feedback. Additionally, we will delve into methods to relay sensation feedback to the user.

\vspace{-0.15cm}
\bibliographystyle{IEEEtran}
\bibliography{root}


\vfill


\end{document}